\pdfoutput=1
\documentclass{article}

\usepackage[preprint]{neurips_2022}

\usepackage[utf8]{inputenc} 
\usepackage[T1]{fontenc}    
\usepackage{hyperref}       
\usepackage{url}            
\usepackage{booktabs}       
\usepackage{amsfonts,bm}       
\usepackage{nicefrac}       
\usepackage{microtype}      
\usepackage{xcolor}         
\usepackage{authblk}
\usepackage{multirow}
\usepackage{wrapfig}
\usepackage{caption}
\usepackage{xspace}
\usepackage{colortbl}
\usepackage{makecell}
\usepackage{graphicx}
\usepackage{enumitem}
\usepackage{subfigure} 
\usepackage{listings}
\usepackage{amsmath}

\usepackage{color}
\definecolor{darkred}{rgb}{0.6,0.0,0.0}
\definecolor{darkgreen}{rgb}{0,0.50,0}
\definecolor{lightblue}{rgb}{0.0,0.42,0.91}
\definecolor{orange}{rgb}{0.99,0.48,0.13}
\definecolor{grass}{rgb}{0.18,0.80,0.18}
\definecolor{pink}{rgb}{0.97,0.15,0.45}

\usepackage{listings}
\definecolor{codegreen}{rgb}{0,0.6,0}
\definecolor{codegray}{rgb}{0.5,0.5,0.5}
\definecolor{codepurple}{rgb}{0.58,0,0.82}
\definecolor{backcolour}{rgb}{0.95,0.95,0.92}
 
\lstdefinestyle{mystyle}{
    backgroundcolor=\color{backcolour},   
    commentstyle=\color{codegreen},
    keywordstyle=\color{magenta},
    numberstyle=\tiny\color{codegray},
    stringstyle=\color{codepurple},
    basicstyle={\small\ttfamily\linespread{1}\selectfont},
    breakatwhitespace=false,         
    breaklines=true,                 
    captionpos=b,                    
    keepspaces=true,                 
    numbers=left,                    
    numbersep=5pt,                  
    showspaces=false,                
    showstringspaces=false,
    showtabs=false,                  
    tabsize=2
}

\newcommand{\sysname}{\textit{PointDP}}

\def\eqref#1{equation~\ref{#1}}

\def\1{\bm{1}}

\def\vtheta{{\bm{\theta}}}

\def\vepsilon{{\bm{\epsilon}}}

\def\vx{{\bm{x}}}

\def\vz{{\bm{z}}}

\DeclareMathAlphabet{\mathsfit}{\encodingdefault}{\sfdefault}{m}{sl}
\SetMathAlphabet{\mathsfit}{bold}{\encodingdefault}{\sfdefault}{bx}{n}

\newcommand{\ie}{\textit{i.e.,}\xspace}
\newcommand{\eg}{\textit{e.g.,}\xspace}
\newcommand{\etal}{\textit{et al.}\xspace}
\newcommand{\wrt}{\textit{w.r.t.}\xspace}

\title{\textit{PointDP}: Diffusion-driven Purification against Adversarial Attacks on 3D Point Cloud Recognition}

\makeatletter
\renewcommand\AB@affilsepx{, \protect\Affilfont}
\makeatother
\author[1]{\bf Jiachen Sun\thanks{\texttt{jiachens@umich.edu}} }
\affil[1]{University of Michigan}
\affil[2]{NVIDIA}
\affil[3]{ASU}
\author[2]{\bf ~Weili Nie}
\author[2]{\bf ~Zhiding Yu}

\author[1]{\bf ~Z. Morley Mao}
\author[2,3]{\bf Chaowei Xiao}

\begin{document}

\maketitle

\begin{abstract}
  3D Point cloud is becoming a critical data representation in many real-world applications like autonomous driving, robotics, and medical imaging. Although the success of deep learning further accelerates the adoption of 3D point cloud in the physical world, deep learning is notorious for its vulnerability to adversarial attacks. In this work, we first identify that the state-of-the-art empirical defense, adversarial training, has a major limitation in applying to 3D point cloud models due to gradient obfuscation. We further propose \sysname, a purification strategy that leverages diffusion models to defend against 3D adversarial attacks. We extensively evaluate \sysname~on six representative 3D point cloud architectures, and leverage 10+ strong and adaptive attacks to demonstrate its lower-bound robustness. Our evaluation shows that \sysname~ achieves significantly better robustness than state-of-the-art purification methods under strong attacks. Results of certified defenses on randomized smoothing combined with \sysname~will be included in the near future.
\end{abstract}

\section{Introduction}
\label{sec:intro}

Point cloud data is emerging as one of the most broadly used representations in 3D computer vision. It is a versatile data format available from various sensors like LiDAR and stereo cameras and computer-aided design (CAD) models, which depicts physical objects by a number of coordinates in the 3D space. Many deep learning-based 3D perception models have been proposed~\cite{wang2015voting,maturana2015voxnet,Riegler2017OctNet,wang2017cnn,qi2017pointnet,choy20194d} and thus realized several safety-critical applications (\eg autonomous driving)~\cite{yin2021center,shi2019pointrcnn,shi2020pv}. 
Although deep learning models~\cite{qi2017pointnet,qi2017pointnet++} have exhibited performance boost on many challenging tasks, extensive studies show that they are notoriously vulnerable to adversarial attacks~\cite{cao2019adversarial,sun2020lidar,xiang2019generating}, where attackers manipulate the input in an imperceptible manner, which will lead to incorrect predictions of the target model. Because of the broad applications of 3D point clouds in safety-critical fields, it is imperative to study the adversarial robustness of point cloud recognition models. 

The manipulation space for 2D adversarial attacks is to change pixel-level numeric values of the input images. Unlike adversarial examples in 2D applications, the flexible representation of 3D point clouds results in an arguably larger attack surface. For example, adversaries could shift and drop existing points~\cite{zheng2019pointcloud}, add new points into the pristine point cloud~\cite{sun2021adversarially}, or even generate new point clouds~\cite{zhou2020lg} to launch attacks. To make attacks less perceptible, different strategies like limiting the number of altered points and constraining the maximal magnitude of shifted points~\cite{sun2021adversarially}. The flexibility of 3D point cloud data formats enables diverse attacks, thus hindering a practical and universal defense design. 

Considering the safety criticalness involved in 3D point cloud applications, various studies have been devoted to advancing the robustness of 3D point cloud recognition models. DUP-Net~\cite{zhou2019dup} and GvG-PointNet++~\cite{dong2020self} pioneered to add statistical outlier removal (SOR) modules as a pre-processing and in-network block, respectively, as mitigation strategies. 
More lately, Sun~\etal~\cite{sun2020adversarial} broke the robustness of DUP-Net and GvG-PointNet++ by specific adaptive attacks. Adversarial training has been acknowledged as the most powerful defense to deliver empirical robustness on PointNet, DGCNN, and PCT~\cite{sun2021adversarially}. Meanwhile, advanced purification strategies like IF-Defense~\cite{wu2020if} and LPC~\cite{li2022robust} leverage more complex module to clean the adversarial point clouds. However, we for the first time demonstrate that standard adversarial training suffers from \textit{gradient obfuscation} in the point cloud recognition models. We also extensively evaluate IF-Defense and LPC to show that their purification strategies are both vulnerable to stronger attacks (\S~\ref{sec:failure}). 

In this work, we further propose \sysname, an adversarial purification method that leverages a diffusion model as a cleanser module to defend against 3D adversaries. Lately, diffusion models have been emerging as dominant generative models~\cite{ho2020denoising,nichol2021improved,dhariwal2021diffusion}, which extend to the 3D space as well~\cite{luo2021diffusion}. Diffusion models have been proven to be effective in defending against attacks in the 2D space~\cite{nie2022diffusion}. Diffusion models take two steps to (i)
diffuse the input data to noise gradually and (ii) reverse the noised data to its origin step by step (\S~\ref{sec:pre}). Besides the high quality of generation, diffusion models add randomness in every step of its process, which could help preventing adaptive adversaries from launching attacks. We rigorously evaluate \sysname~with six representative point cloud models and sixteen attacks. \sysname~on average achieves 75.9\% robust accuracy while maintaining similar clean accuracy to the original models.

In a nutshell, our contributions are summarized as \textit{two-fold}:
\begin{itemize}[leftmargin=*] 
\setlength{\itemsep}{1pt}
\setlength{\parskip}{2pt}
\item We for the first time demonstrate that standard adversarial training~\cite{madry2017towards,sun2021adversarially}, the most longstanding defense in the 2D space, has a major limitation of application in 3D point cloud models due to architecture designs. We leverage black-box attacks to demonstrate our claim that drop adversarially trained models' robust accuracy to $\sim$10\%.

\item We propose \sysname~that leverage diffusion models to purify adversarial examples. We conduct extensive and rigorous evaluation on six representative models with numerous attacks to comprehensively understand the robustness of \sysname. Our evaluation shows that \sysname~outperforms state-of-the-arts purification methods, IF-Defense~\cite{wu2020if} and LPC~\cite{li2022robust} by 12.6\% and 40.3\% on average, respectively. 
\end{itemize}

\section{\sysname: Diffusion-driven Purification against 3D Adversaries}

We first introduce the preliminaries of diffusion models and then propose \sysname~that first introduces noise to the adversarial 3D point clouds, followed by the forward process of diffusion models to get diffused point clouds. Purified point clouds are recovered through the reverse process (\S-\ref{sec:pointdp}). Next, we follow~\cite{nie2022diffusion} to apply the adjoint method to backward propagate through SDE for efficient gradient evaluation with strong adaptive attacks (\S~\ref{sec:adaptive}).

\subsection{Preliminaries}
\label{sec:pre}

In this section, we briefly review the background of diffusion models in 3D vision tasks. As mentioned in~\S~\ref{sec:intro}, diffusion models involve the forward and reverse processes. 

Given a clean point cloud sampled from the unknown data
distribution $\vx(0) \sim q(\vx)$, the forward process of the diffusion model leverages a fixed Markov chain to gradually add Gaussian noise to the clean point cloud $\vx_0$ over a pre-defined $T$ time steps, resulting in a number of noisy point clouds $\vx(1)$, $\vx(2)$, ..., $\vx(T)$. Mathematically, the forward process is defined as
\begin{equation}
\begin{split}
    q(\vx(1:T)|\vx(0)) &:= \prod_{t=1}^{T}q(\vx(t)|\vx(t-1)),\\
    q(\vx(t)|\vx(t-1)) &:= \mathcal{N}(\vx(t);\sqrt{1-\beta(t)}\vx(t-1),\beta(t)\mathbf{I})
\end{split}
\label{eq:1}
\end{equation}
where $\beta(t)$ is a scheduling function of the added Gaussian noise (\eg $\beta(t) = \beta_{\min} + (\beta_{\max} - \beta_{\min})t$). 

The reverse process, in contrast, is trained to recover the diffused point cloud in an iterative manner. 3D Point clouds have less semantics than 2D images due to the lack of texture information. Therefore, point cloud diffusion models leverage a separate encoder $e$ to as a latent feature $z_\vx = e(\vx)$ as a condition to help recover the clean point cloud. 
\begin{equation}
\begin{split}
    p_\vtheta(\vx(0:T)|\vz) &:= p(\vx(T))\prod_{t=1}^{T}p_\vtheta(\vx(t-1)|\vx(t),\vz), \\ p_\vtheta(\vx(t-1)|\vx(t),\vz) &:= \mathcal{N}(\vx(t-1)|\mu_\vtheta(\vx(t),t,\vz),\beta(t)\mathbf{I})
\end{split}
\label{eq:2}
\end{equation}
where $\mu_\vtheta$ denotes the approximated mean value parameterized by a neural network, and $\vz = e(\vx_a)$. The training objective is to learn the variational bound of the negative log-likelihood~\cite{luo2021diffusion}. In practice, we jointly train the encoder $e$ with $\mu_\vtheta$. 
Essentially, the sampling process is similar to the DDPM model~\cite{dhariwal2021diffusion}:
\begin{equation}
    \vx(t-1) = \frac{1}{\sqrt{\alpha(t)}}(\vx(t)-\frac{1-\alpha(t)}{\sqrt{1-\overline{\alpha(t)}}}\vepsilon_\vtheta(\vx(t),t,\vz))
\label{eq:reverse}
\end{equation}
where $\alpha(t) = \prod_{s=1}^t \alpha(s)$. Point cloud diffusion models have recently achieved SOTA performance on generating and autoencoding 3D point clouds, which provides us with opportunities for adversarial point cloud purification.  

\begin{figure}[t]
    \centering
    \includegraphics[width=\linewidth]{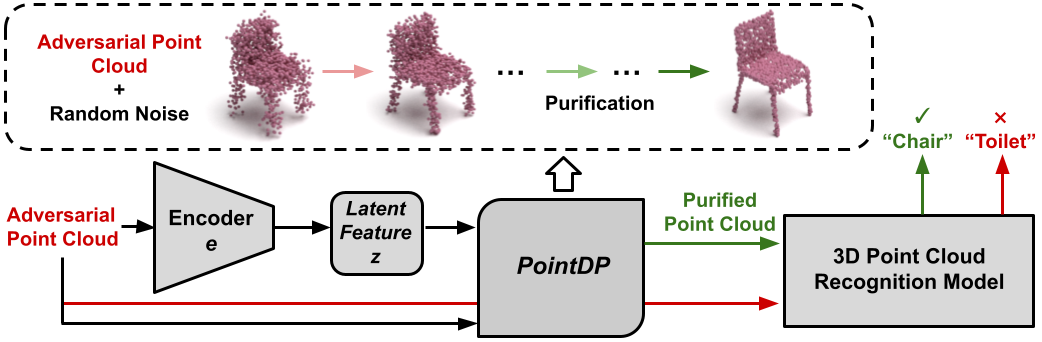}
    \caption{Illustration of \sysname. We leverage~\cite{luo2021diffusion} as the diffusion model in our study.}
    \label{fig:overall}
\end{figure}

\subsection{Design of \sysname}
\label{sec:pointdp}

Figure~\ref{fig:overall} illustrates the pipeline of \sysname. Nie~\etal~\cite{nie2022diffusion} have shown that diffusion-driven purification is able to remove the adversarial effect for 2D images. As mentioned in \S~\ref{sec:pre}, conditional diffusion models were proposed in the 3D point cloud space. Specifically, we use the design in~\cite{luo2021diffusion} as the base model for the purification process in our study. Note that we do not aiming at designing new point cloud diffusion models, but instead propose a novel purification pipeline along with rigorous evaluation as our main contributions.

Let $\vx_a$ be an adversarial example \wrt the pristine classifier $f$, we initialize the input of the forward diffusion process as $\vx_a$, \ie $\vx(t=0)=\vx_a$. The forward diffusion process can be solved by Equation~\ref{eq:3} from $t=0$ to $t=t^*$
\begin{equation}
    \vx(t^*) = \sqrt{\alpha(t^*)}\vx_a + \sqrt{1-\alpha(t^*)}\vepsilon
\label{eq:3}
\end{equation}
where $\alpha(t) = e^{-\int_{0}^{t}\beta(s)ds}$, $\vepsilon \sim \mathcal{N}(0,\mathbf{I})$. We leverage Equation~\ref{eq:reverse} to recover the clean point clouds.
Equivalently, the reverse can be also solved by the SDE solver in~\cite{nie2022diffusion}, noted as: $\mathbf{sdeint}$:
\begin{equation}
    \hat{\vx(0)} = \mathbf{sdeint}(\vx(t^*),f_{rev},g_{rev},w,t^*,0)
    \label{eq:sde}
\end{equation}
where the six inputs are initial value, drift coefficient, diffusion coefficient, Wiener process, initial time, and end time~\cite{nie2022diffusion}.
\begin{equation}
\begin{split}
    f_{rev}(\vx,t) &= -\frac{1}{2}\beta(t)[\vx+2s_\vtheta(\vx,t)] \\
    g_{rev}(t) &= \sqrt{\beta(t)}
\end{split}
\end{equation}
where score function $s_\vtheta$ is derived from $s_\vtheta(\vx,t) = -\frac{1}{\sqrt{1-\alpha(t)}}\vepsilon_\vtheta (\vx(t),t,\vz)$~\cite{ho2020denoising}.

Besides, the hyper-parameter $t*$ and $T$ trades off the denoising performance and efficiency. We empirically choose $t^*=10$ and $T=200$ in our study, which has shown satisfactory results in our evaluation (\S~\ref{sec:eval}).

\subsection{Adpative Attacks on \sysname}
\label{sec:adaptive}
\sysname~is a pre-processing module that purifies the adversarial perturbations.~\cite{athalye2018obfuscated} have shown that input transformation-based methods can be broken by specifically designed attacks. Therefore, it is essential to model the adaptive attacks on \sysname~to demonstrate its lower-bound adversarial robustness. We thus formulate two types of adaptive attacks on \sysname. 

\textbf{Attack on Latent Feature}. As \sysname~utilizes conditional diffusion models for adversarial purification, the latent feature $z$ is a good candidate for adversaries to launch attacks. Concretely, adversaries can set the goal to maximize some distance metric $\mathcal{D}$ between the latent feature of the optimized adversarial examples and the oracle latent feature of clean inputs $\vz_\text{oracle}$. Without loss of generality, the adaptive attacks can be formulated as:
\begin{equation}
    \vx_{s+1} = \mathrm{Proj}_{\vx + \mathcal{S}}(\vx_{s} + \alpha \cdot \mathrm{norm}(\nabla_{\vx_{s}} \mathcal{D}(e(\vx_s),\vz_\text{oracle}))),
    \label{eq:pgd}
\end{equation}
where $\vx_s$ denotes the adversarial examples from the $s$-th step, $\mathrm{Proj}$ is the function to project the adversarial examples to the pre-defined space $\mathcal{S}$, and $\alpha$ is the attack step size. We choose two distance metrics in our study, where the first one is the KL divergence~\cite{goldberger2003efficient} and the other is the the $\ell_1$ norm distance. In our evaluation (\S~\ref{sec:eval}), we report the lowest accuracy achieved under attacks with two distance metrics.

\textbf{Attack Using BPDA}. We follow~\cite{nie2022diffusion} to formulate the adaptive attack as an augmented SDE process. We re-state the attack formulation as below.
For the SDE in Equation~\ref{eq:sde}, the augmented SDE that computes the gradient $\frac{\partial \mathcal{L}}{\partial {\vx}(t^*)}$ of backward propagating through it is given by:
\begin{align}\label{aug_sdeint}
\renewcommand\arraystretch{1.5}
    \! \begin{pmatrix}
        {\vx}(t^*) \\
           \frac{\partial \mathcal{L}}{\partial {\vx}(t^*)} 
    \end{pmatrix}
    \! = \! \texttt{sdeint} \! \left( 
        \! \begin{pmatrix}
        \hat{\vx}(0) \\ 
           \frac{\partial \mathcal{L}}{\partial \hat{\vx}(0)} 
    \end{pmatrix}\!,
    \tilde{f}, \tilde{g}, \tilde{w}, 0, t^*
    \right)
\end{align}
where $\frac{\partial \mathcal{L}}{\partial \hat{\vx}(0)}$ is the gradient of the objective $\mathcal{L}$ w.r.t. the output $\hat{\vx}(0)$ of the SDE in Equatrion~\ref{eq:sde}), and
\begin{align*}
    \begin{split}
        \tilde{f}([\vx; \vz], t) &=
        \renewcommand\arraystretch{1.5}
        \begin{pmatrix}
            f_{\text{rev}}(\vx, t) \\
            {\frac{\partial f_{\text{rev}}(\vx, t) }{\partial \vx}} \vz
        \end{pmatrix}  \\
        \renewcommand\arraystretch{1.5}
        \tilde{g}(t) &=
        \begin{pmatrix}
            -g_{\text{rev}}(t) \mathbf{1}_d \\
            \mathbf{0}_d
        \end{pmatrix} \\
        \renewcommand\arraystretch{1.5}
        \tilde{w}(t) &=
        \begin{pmatrix}
            -w(1-t) \\
            -w(1-t)
        \end{pmatrix}
    \end{split}
\end{align*}
where $\mathbf{1}_d$ and $\mathbf{0}_d$ denote the $d$-dimensional vectors of all ones and all zeros, respectively. Nie~\etal~\cite{nie2022diffusion} have demonstrated that such approximation align well with the true gradient value. Therefore, we leverage this adaptive attack formulation for our evaluation.

\section{Related Work}
\label{sec:related}

In this section, we review the current progress of deep learning, adversarial attacks, and defenses for 3D point cloud recognition tasks.

\subsection{Deep Learning on 3D Point Cloud Recognition}

2D computer vision has achieved stellar progress on architectural designs of convolutional neural networks~\cite{he2016deep}, followed by vision transformers~\cite{dosovitskiy2020image}. However, there is currently no consensus on the architecture of 3D perception models since there is no standard data format for 3D perception~\cite{sun2022benchmarking}. As raw data from both 3D scanners and triangular meshes can be efficiently transformed into point clouds, they are becoming the most often utilized data format in 3D perception. 3D networks at the early stage use dense voxel grids for perception~\cite{wang2015voting,maturana2015voxnet,DeepSlidingShapes,tchapmi_segcloud_3dv17}, which discretize a point cloud to voxel cells for classification, segmentation, and object detection.
PointNet pioneered to leverage global pooling help achieve memory-efficient permutation invariance in an end-to-end manner. PointNet++~\cite{qi2017pointnet++} and DGCNN~\cite{wang2019dynamic} followed up to add sophisticated local clustering operations to advance the performance. Sparse tensors are the other direction in 3D network designs~\cite{SubmanifoldSparseConvNet,choy20194d} to use 3D convolutions to improve 3D perception performance. PointCNN and RSCNN reformed the classic pyramid CNN to improve the local feature generation~\cite{li2018pointcnn,liu2019relation}. PointConv and KPConv designed new convolution operation for point cloud learning~\cite{wu2019pointconv,thomas2019kpconv}. PointTransformer and PCT advanced self-attention blocks in the 3D space and achieved 
good performance~\cite{zhao2021point,guo2020pct}. Various novel local clustering operations~\cite{xiang2021walk,ma2022rethinking} also show enhancements on the clean performance.
In this work, we focus on PointNet, PointNet++, DGCNN, PCT, CurveNet, and PointMLP as our evaluation backbones since they are representative and widely used and achieve state-of-the-art results in point cloud recognition~\cite{mn40}. 

\subsection{Adversarial Attacks and Defenses}
\label{sec:related_3d_attack}

Adversarial attacks have become the main obstacle that hinder deep learning models from real-world deployments, especially in safety-critical applications~\cite{eykholt2018robust,sun2020lidar,cao2019adversarial,zhang2021emp,Zhang_2022_CVPR}. There are a lot of adversarial attacks proposed in the 2D space to break the various vision models~\cite{carlini2017towards,xiao2018generating,yang2020patchattack, xie2017adversarial,huang2019universal,huang2020universal,xiao2018spatially,sun2021certified}. To fill this gap between standard and robust accuracies, many mitigation solutions have been studied and presented to improve the robustness against adversarial attacks~\cite{yang2019me,xu2017feature,bafna2018thwarting,papernot2016distillation,meng2017magnet,zhang2019towards,xiao2018characterizing,zhang2020robust,xiao2019advit} and 3D domains~\cite{dong2020self,zhou2019dup,sun2020adversarial}. However, most of them including adding randomization~\cite{liu2019extending,dhillon2018stochastic,dong2020self}, model distillation~\cite{papernot2016distillation}, adversarial detection~\cite{meng2017magnet}, and input transformation~\cite{yang2019me,xu2017feature,papernot2017extending,bafna2018thwarting,zhou2019dup} have been compromised by adaptive attacks~\cite{sun2020adversarial,tramer2020adaptive,athalye2018obfuscated}. Adversarial training (AT)~\cite{madry2017towards,goodfellow2014explaining,wong2020fast,shafahi2019adversarial}, in contrast, delivered a more longstanding mitigation strategy~\cite{xie2020smooth,Xie2020Intriguing,zhang2019theoretically}. However, the robust accuracy achieved by AT is still not satisfactory enough to be used in practice.
Most recently, Nie~\etal proposed DiffPure~\cite{nie2022diffusion} that leverages diffusion models to defend against adversarial attacks, and following-up studies to extend it to certified defenses~\cite{carlini2022certified}.

Adversarial attacks and defenses also extend to 3D point clouds. Xiang~\etal~\cite{xiao2018generating} first demonstrated that point cloud recognition models are vulnerable to adversarial attacks. They also introduced different threat models like point shifting and point adding attacks. Wen~\etal~\cite{wen2019geometry} enhanced the loss function in C\&W attack to achieve attacks with smaller perturbations and Hamdi~\etal~\cite{hamdi2020advpc} presented transferable black-box attacks on point cloud recognition. Zhou~\etal~\cite{zhou2019dup} and Dong\etal~\cite{dong2020self} proposed to purify the adversarial point clouds by input transformation and adversarial detection. However, these methods have been successfully by~\cite{sun2020adversarial} through adaptive attacks. Moreover, Liu~\etal~\cite{liu2019extending} made a preliminary investigation on extending countermeasures in the 2D space to defend against simple attacks like FGSM~\cite{goodfellow2014explaining} on point cloud data. Sun~\etal~\cite{sun2021adversarially} conducted a more thorough study on the application of self-supervised learning in adversarial training for 3D point clodu recognition. Besides adversarial training, advanced purification methods IF-Defense~\cite{wu2020if} and LPC~\cite{li2022robust} were proposed to transform the adversarial examples to the clean manifold. In this work, we present \sysname, that utilizes 3D diffusion models to purify adversarial point clouds. We also demonstrate that standard adversarial training suffer from strong black-box attacks and SOTA purification methods (\ie IF-Defense and LPC) are vulnerable to PGD-styled adversaries (\S~\ref{sec:failure}).
\section{Experiments and Results}
\label{sec:eval}

In this section, we first introduce our experimental setups (\S~\ref{sec:setup}). We then present the standard robustness evaluation of \sysname (\S~\ref{sec:exp_dp}). We next show that how the SOTA adversarial training and adversarial purification methods fail under various strong attacks (\S~\ref{sec:failure}). We finally conduct stress test on \sysname~to show its actual robustness under various stronger adaptive attacks (\S~\ref{sec:unseen}).

\subsection{Experimental Setups}
\label{sec:setup}

\textbf{Datasets and Network Architectures}. We conduct all the experiments on the widely used ModelNet40 point cloud classification benchmark~\cite{wu20153d}, consisting of 12,311 CAD models from 40 artificial object categories. We adopt the official split with 9,843 samples for training and 2,468 for testing. We also uniformly sample 1024 points from the surface of each object and normalize them into an edge-length-2 cube, following most of the prior arts~\cite{qi2017pointnet}. As mentioned before, there are various backbones for 3D point cloud recognition in the literature. To demonstrate the universality of \sysname, we select six representative model architectures including PointNet~\cite{qi2017pointnet}, PointNet++~\cite{qi2017pointnet++}, DGCNN~\cite{wang2019dynamic}, PCT~\cite{guo2020pct}, CurveNet~\cite{xiang2021walk}, and PointMLP~\cite{ma2022rethinking}. These backbones either have representative designs (\textit{e.g.}, Transformer and MLP) or achieve SOTA performance on the ModelNet40 benchmark. 

\textbf{Adversarial Attacks}. As briefly described in \S~\ref{sec:related_3d_attack}, adversarial attacks could be roughly categorized into C\&W- and PGD-styled attacks. C\&W attacks involves the perturbation magnitude into the \textit{objective} term of the optimization procedure, while PGD attacks set the perturbation magnitude as a firm \textit{constraint} in the optimization procedure. Moreover, adversarial attacks by $\ell_p$ norm as the distance metric for the perturbation. Although a number of attacks measure Chamfer and Handoff ``distances'' in 3D point cloud~\cite{xiang2019generating}, they are not formal distance metrics as they do not satisfy the triangular inequality. Therefore, we still leverage $\ell_2$ and $\ell_\infty$, following most defense studies in both 2D and 3D vision tasks~\cite{carlini2017towards,sun2021adversarially}. We also have designed adaptive attacks on our proposed method~\S~\ref{sec:adaptive}.
Besides naive C\&W and PGD attacks, we leverage specific attacks designed to break the robustness of point cloud recognition such as $k$NN~\cite{tsai2020robust} and AdvPC~\cite{hamdi2020advpc}. We also apply strong adaptive AutoAttack~\cite{croce2020reliable} (\textit{i.e.}, APGD) in our evaluation. Moreover, we use SPSA~\cite{uesato2018adversarial} and Nattack~\cite{li2019nattack} as black-box adversaries, followed by the suggestion of Carlini~\etal~\cite{carlini2019evaluating}. We also leverage EOT-AutoAttack. Point adding (PA) and dropping/detaching (PD) attacks are also evaluated in our study, followed by the setups in~\cite{sun2021adversarially}. We set the attack steps to 200 to maximize the adversarial capability and follow the settings in~\cite{sun2021adversarially} for other attack parameters by default.

\textbf{Evaluation Metrics}. We leverage two main metrics to evaluate the performance of our defense proposal, which are \textit{standard} and \textit{robust} accuracy. The standard accuracy measures the performance of the defense method on clean data, which is evaluated on the whole test set from ModelNet40. The robust accuracy measures the performance on adversarial examples generated by different attacks. Because of the high computational cost of applying \textit{adaptive} and \textit{black-box} attacks to our method, we evaluate robust accuracy for our defense on a fixed subset of 128 point clouds randomly sampled from the test set. Notably, robust accuracies of most baselines do not change much on the sampled subset, compared to the whole test set. We evaluate the robust accuracy on the whole test set for other adversarial attacks with acceptable overhead (\textit{e.g.}, C\&W and PGD attacks).

\begin{table}[ht]
\renewcommand\arraystretch{1.1}
\setlength\tabcolsep{14pt}
  \caption{Evaluation Results of Plain Model on PA and PD (Accuracy \%). Models under other attacks mostly have \textbf{0\%} accuracy, so we do not present them here.}
  \label{tb:plain}
  \centering
  \resizebox{0.9\linewidth}{!}{
  
    \begin{tabular}{c|cccccc}
\noalign{\global\arrayrulewidth1pt}\hline\noalign{\global\arrayrulewidth0.2pt}
&PointNet &PointNet++ &DGCNN &PCT &CurveNet &PointMLP  \\
\hline
\rowcolor{gray!20} 
    None &90.1 &92.8 &92.5 &92.8 &93.2 &93.5\\
\hline
   PA &44.1 &19.9 &35.1 &20.8 &48.9 &7.2\\
   PD &33.3 &69.8 &64.5 &53.0 &72.6 &71.1\\
\noalign{\global\arrayrulewidth1pt}\hline\noalign{\global\arrayrulewidth0.2pt}
\end{tabular}
  }
\end{table}

\textbf{Baseline}. Without any defense applied to the original recognition models, the robust accuracy is mostly \textbf{0} for all models under $\ell_2$ and $\ell_\infty$ based attacks. DGCNN exceptionally achieves 64\% on $\ell_2$-based PGD, AutoAttack, respectively, due to its dynamic clustering design, which adaptively discards outlier points. PA and PD are two weaker attacks and Table~\ref{tb:plain} presents the robust accuracy against these two attacks.

\subsection{Experiment Results of \sysname}
\label{sec:exp_dp}

In this section, we first present the evaluation results of \sysname under attacks on the plain models.
We train the diffusion and 3D point cloud recognition models in a sequential order. 
Table~\ref{tb:results_pd} presents the detailed results of \sysname~against attacks on six models. We find that \sysname~overall achieves satisfactory results across all models and attacks. The average robust accuracy against adversarial attacks is above 75\%. We observe a drop on the clean accuracy for the chosen models, which is expected. As mentioned before, diffusion models for 3D point cloud is a more difficult task than 2D image diffusion, which may lead to partial semantic loss. The average drop of standard accuracy is 4.9\%. We find that DGCNN still achieves the best robustness combined with \sysname, which has a 79.9\% of robust accuracy. We further compare the performance of \sysname with adversarial training, IF-Defense, and LPC in the next section.

\begin{table}[ht]
\renewcommand\arraystretch{1.1}
\setlength\tabcolsep{12pt}
  \caption{Evaluation Results of Adversarial Attacks on \sysname~(Accuracy \%).}
  \label{tb:results_pd}
  \centering
  \resizebox{1.\linewidth}{!}{
  
    \begin{tabular}{l|c|cccccc}
\noalign{\global\arrayrulewidth1pt}\hline\noalign{\global\arrayrulewidth0.2pt}
& &PointNet &PointNet++ &DGCNN &PCT &CurveNet &PointMLP  \\
\hline
\rowcolor{gray!20} 
   & None & 86.8 & 87.9 & 86.9 & 87.0 & 88.0 &88.2\\
\hline
\multirow{4}{*}{\makecell[c]{$\ell_\infty$ \\ $\epsilon=0.05$}} &C\&W &77.9 &78.6 &78.9 &76.8 &73.1 &76.2\\
    & PGD &78.1 &80.6 &80.3 &77.2 &74.8 &79.8\\
    & AdvPC &69.7 &76.6 &79.1 &79.4 &72.6 &75.2\\
    & PA & 82.1 &85.1 &84.8 &85.5 &86.3 &85.8\\
\hline
\multirow{4}{*}{\makecell[c]{$\ell_2$ \\ $\epsilon=1.25$}} &C\&W &82.4 &82.9 &81.9 &80.9 &81.5 &82.6\\
    & PGD &80.1 &75.0 &74.6 &72.0 &71.7 &76.4\\
    & AdvPC & 69.1 & 76.3 & 79.0 &74.2 & 74.1 &75.6\\
    & $k$NN & 83.5 & 82.9 &83.3 &82.3 &81.5 &83.1\\
\hline
\makecell[c]{$\ell_0$ \\ $\epsilon=200$} & PD &68.9 &74.1 &77.3 &76.3 &76.8 &77.4 \\
\noalign{\global\arrayrulewidth1pt}\hline\noalign{\global\arrayrulewidth0.2pt}
\end{tabular}
  }
\end{table}

\subsection{Failure of State-of-the-Art Defenses}
\label{sec:failure}

\begin{figure}[t]
    \centering
    \includegraphics[width=\linewidth, height=3.5cm]{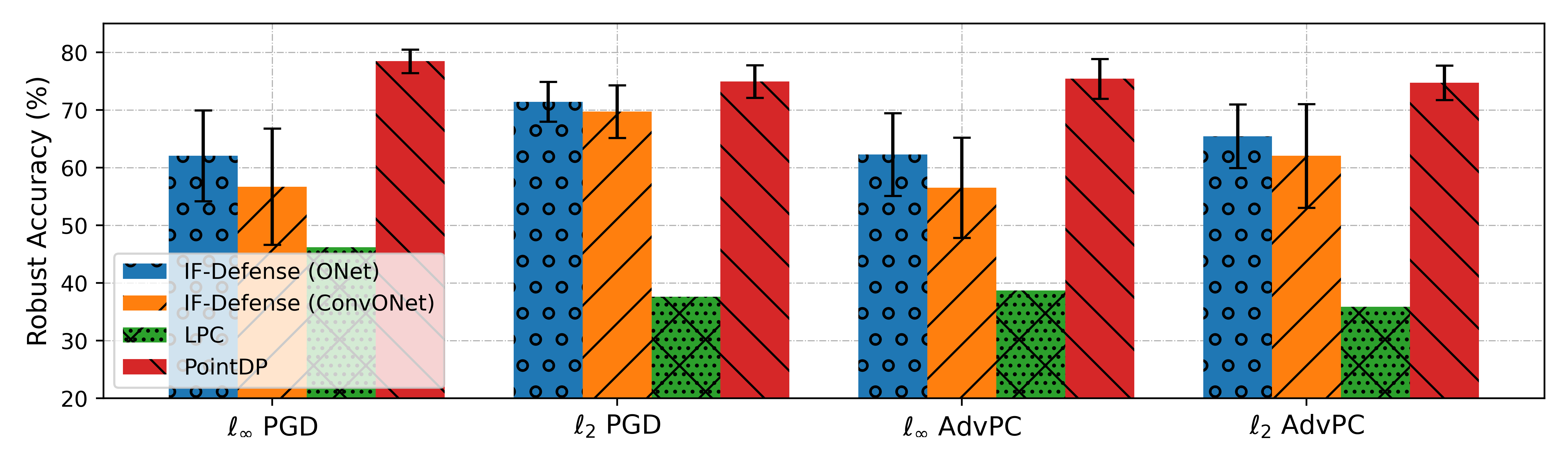}
    \caption{Compare among SOTA Adversarial Purification Strategies (\ie IF-Defense~\cite{wu2020if}, LPC~\cite{li2022robust}, and \sysname). The results of IF-Defense and \sysname~are averaged from six models.}
    \label{fig:compare}
\end{figure}

\begin{figure}[t]
\begin{minipage}[t]{\linewidth}
\begin{lstlisting}
def knn(x, k):
    inner = -2*torch.matmul(x.transpose(2, 1), x)
    xx = torch.sum(x**2, dim=1, keepdim=True)
    pairwise_distance = -xx - inner - xx.transpose(2, 1)
    idx = pairwise_distance.topk(k=k, dim=-1)[1]   
    # (batch_size, num_points, k)
    return idx
    
def get_graph_feature(x, k): 
    # x's shape is (batch_size, num_dims, num_points)
    idx = knn(x, k=k)   # (batch_size, num_points, k)
    ...... # shape transformation here
    feature = x.view(batch_size*num_points, -1)[idx, :] 
    # idx is used as index to select features, leading to gradient obfuscation
    ...... 
    return feature

# forward function for EdgeConv
def forward(self, x):
    ......
    x = get_graph_feature(x, k=self.k) # kNN hyperparameter
    x = self.conv1(x) # convolution 
    x1 = x.max(dim=-1, keepdim=False)[0] # max pooling
    ......
\end{lstlisting}
\end{minipage}
\caption{PyTorch~\cite{paszke2019pytorch}-Style Code Snippet of EdgeConv~\cite{wang2019dynamic} in Point Cloud Recognition Models. Adversarial training fails since the $k$NN layers leverage the top-$k$ function where the gradient propagate to the index, resulting in gradient obfuscation.}
\label{fig:at_failure}
\end{figure}

In this section, we demonstrate how lately proposed defense solutions fail when encountered with stronger (adaptive) adversarial attacks on 3D point cloud recognition models. 

\begin{wraptable}{r}{7cm}
\renewcommand\arraystretch{1.1}
\setlength\tabcolsep{10pt}
  \caption{Evaluation Results of Standard Adversarial Training (Accuracy \%) $\ell_\infty$ $\epsilon=0.05$.}
  \label{tb:at}
  \centering
  \resizebox{0.9\linewidth}{!}{
  
    \begin{tabular}{c|ccc}
\noalign{\global\arrayrulewidth1pt}\hline\noalign{\global\arrayrulewidth0.2pt}
&PointNet &DGCNN &PCT  \\
\hline
\rowcolor{gray!20} 
    None &87.8  &90.6 &89.7 \\
\hline
    PGD &52.1 &67.4 &51.3 \\
    AutoAttack &40.5 &56.4 &47.2\\
\hline
    SPSA &56.7 &\textcolor{red}{7.8} &\textcolor{red}{11.4}\\
    Nattack &55.1 &\textcolor{red}{5.4} &\textcolor{red}{6.5}\\
\noalign{\global\arrayrulewidth1pt}\hline\noalign{\global\arrayrulewidth0.2pt}
\end{tabular}
  }
\end{wraptable}

\textbf{Adversarial training} (AT) has been applied to PointNet, DGCNN, and PCT with the help of self-supervised learning~\cite{sun2021adversarially} that achieves satisfactory robustness. Such observations are consistent with the performance of AT for 2D perception models. However, we find that AT is, in fact, a weak defense solution in 3D perception models. First, as acknowledged by~\cite{sun2021adversarially}, point cloud models (\eg PointNet++ and CurveNet) often leverage different sampling strategies to select anchor points, like furthest point sampling (FPS). Such sampling involves high randomness. AT either cannot converge with different random seeds in each iteration or overfits to a single random seed. Therefore, AT cannot fit these models. Moreover, we discover that the $k$NN layers will cause severe \textit{gradient obfuscation} in point cloud models as well. Different from 2D models that are almost fully differentiable, except for the max pooling layer. As shown in Figure~\ref{fig:at_failure}, $k$NN essentially applies top-$k$. Therefore, gradient backward propagation through $k$NN layers is indexing, which is not non-smooth. The heavy usage of $k$NN layers in DGCNN and PCT will drastically hinder the gradient flow. As mentioned in \S~\ref{sec:setup}, we exploit black-box SPSA and Nattack to validate our findings. Table~\ref{tb:at} presents the results of AT. SPSA and Nattack can greatly lower the average robust accuracy (7.8\%) than white-box attacks (55.6\%) on DGCNN and PCT, which confirms the effect of gradient obfuscation. PointNet, however, achieves better robustness under black-box attacks because it only has one max pooling layer and does not employ $k$NN layers. 

\begin{table}[ht]
\renewcommand\arraystretch{1.1}
\setlength\tabcolsep{12pt}
  \caption{Evaluation Results of Adversarial Attacks on IF-Defense (Accuracy \%).}
  \label{tb:if}
  \centering
  \resizebox{1.\linewidth}{!}{
  
    \begin{tabular}{l|c|cccccc}
\noalign{\global\arrayrulewidth1pt}\hline\noalign{\global\arrayrulewidth0.2pt}
& &PointNet &PointNet++ &DGCNN &PCT &CurveNet &PointMLP  \\
\hline
\rowcolor{gray!20} 
   ONet & None &90.0 &92.8 &92.4 &92.8 &93.1 &93.5 \\
\hline
\multirow{2}{*}{\makecell[c]{$\ell_\infty$ \\ $\epsilon=0.05$}}
    & PGD &69.9 &74.0 &61.0 &54.1 &51.9 &61.6\\
    & AdvPC &69.4 &72.8 &61.6 &53.9 &53.6 &62.5\\
\hline
\multirow{2}{*}{\makecell[c]{$\ell_2$ \\ $\epsilon=1.25$}}
    & PGD &74.2 &77.5 &70.5 &67.2 &68.7 &70.5\\
    & AdvPC &69.0 &72.9 &63.0 &64.5 &55.4 &67.9\\
\hline
\rowcolor{gray!20} 
   ConvONet & None &90.1 &92.8 &92.5 &92.8 &93.2 &93.5 \\
\hline
\multirow{2}{*}{\makecell[c]{$\ell_\infty$ \\ $\epsilon=0.05$}}
    & PGD &66.4 &73.2 &52.9 &46.8 &45.3 &55.7\\
    & AdvPC &63.7 &71.2 &55.5 &47.2 &46.7 &55.0\\
\hline
\multirow{2}{*}{\makecell[c]{$\ell_2$ \\ $\epsilon=1.25$}}
    & PGD & 72.2 & 76.7 &69.8 &65.6 &62.7 &71.4\\
    & AdvPC &63.4 &74.3 &56.6 &59.8 &47.2 &71.0\\
\noalign{\global\arrayrulewidth1pt}\hline\noalign{\global\arrayrulewidth0.2pt}
\end{tabular}
  }
\end{table}

\textbf{Existing purification-based defenses} against 3D adversarial point clouds mainly leverage C\&W-styled attacks in their evaluation. C\&W attacks utilize the method of Lagrange multipliers to find tractable adversarial examples while minimizing the magnitudes of the perturbation. From the perspective of adversary, such attacks are desirable due to their stealthiness, while this does not hold from a defensive view. Defense methods should be evaluated against strong adaptive attacks~\cite{carlini2019evaluating}. IF-Defense and LPC are the SOTA adversarial purification methods for 3D point cloud models. We leverage PGD and AdvPC attacks, which assign constant adversarial budget in the adversarial optimization stage. We follow the original setups of IF-Defense and LPC in our study. Such evaluation is stronger than C\&W attacks, while we note that they are not strict adaptive attacks since the adversarial target is still the classifier itself. Similar to \sysname~, IF-Defense can be pre-pended to any point cloud classifier, but LPC uses a specific backbone. Table~\ref{tb:if} presents the detailed evaluation results of IF-Defense under various settings and attacks. We find that \sysname~achieves much better robustness than IF-Defense, which is on average an 12.6\% improvements. However, IF-Defense achieves slightly higher clean accuracy (4.9\%). This is because IF-Defense leverages SOR to smooth the point cloud~\cite{zhou2019dup}. However, such an operation has been demonstrated to be vulnerable~\cite{sun2020adversarial}. With specific adaptive attacks, there will be a even larger drop of robust accuracy for IF-Defense.

Figure~\ref{fig:compare} shows the comparison among \sysname~and existing methods. \sysname~overall achieves the best performance than prior arts, which are 12.6\% and 40.3\% improvements than IF-Defense and LPC, respectively. We find that even without adaptive attacks, adversaries with constant budgets can already hurt the robust accuracy by a significant margin. This suggests that IF-Defense and LPC fail to deliver strong robustness to 3D point cloud recognition models. Especially, LPC appears in the proceedings of CVPR 2022, but actually achieves trivial robustness, indicating that a rigorous benchmarking is highly required in this community.

\subsection{Defense against Adaptive Threats}
\label{sec:unseen}

We have so far illustrated that state-of-the-art defenses can be easily broken by (adaptive) adversarial attacks and \sysname~consistently achieves the best robustness. In this section, we further extensively evaluate the robustness of \sysname~on even stronger adaptive attacks to demonstrate the actual robustness realized by \sysname. As mentions in \S~\ref{sec:setup}, we leverage two types of adaptive attacks in our study, and Table~\ref{tb:adaptive} presents their results. We also leverage black-box SPSA and Nattack to validate our results. We find that BPDA-PGD the strongest adaptive attacks, which align well with previous study on 2D diffusion-driven purification~\cite{nie2022diffusion}. Even though with strong adaptive attacks, \sysname~still achieves much better robustness. Besides, black-box attacks are much less effective. Although we admit that \sysname~still relies on gradient obfuscation, the extremely high randomness will hinder the black-box adversaries finding correct gradients. 

\begin{table}[ht]
\renewcommand\arraystretch{1.1}
\setlength\tabcolsep{10pt}
  \caption{Evaluation Results of \textbf{Strong Adaptive} Attacks on \sysname~(Accuracy \%).}
  \label{tb:adaptive}
  \centering
  \resizebox{1.\linewidth}{!}{
  
    \begin{tabular}{l|c|cccccc}
\noalign{\global\arrayrulewidth1pt}\hline\noalign{\global\arrayrulewidth0.2pt}
& &PointNet &PointNet++ &DGCNN &PCT &CurveNet &PointMLP  \\
\hline
\rowcolor{gray!20} 
   & None & 86.8 & 87.9 & 86.9 & 87.0 & 88.0 &88.2\\
\hline
\multirow{7}{*}{\makecell[c]{$\ell_\infty$ \\ $\epsilon=0.05$}}
    & BPDA-PGD &77.1 &78.6 &79.2 &76.1 &73.9 &77.7\\
    & EOT-AutoAttack &78.0 &79.9 &79.1 &76.5 &75.9 &78.9\\
    & PGD & 80.8 &80.7 &82.9 &82.5 &80.8 &79.9\\
    & AdvPC &69.9 &76.8 &79.4 &79.8 &72.9 &75.4\\
    & SPSA &76.6 &78.9 &74.9 &78.5 &76.4 &80.9\\
    & Nattack &75.2 &77.9 &74.4 &78.0 &76.1 &78.9\\
    & PA & 81.7 &84.7 &84.1 &84.5 &84.8 &85.2\\
\hline
\multirow{6}{*}{\makecell[c]{$\ell_2$ \\ $\epsilon=1.25$}}
    & BPDA-PGD &78.9 &73.3 &73.3 &71.2 &70.7 &75.1\\
    & EOT-AutoAttack &79.6 &74.4 &74.2 &71.3 &71.3 &75.9\\
    & PGD &86.1 &87.5 &82.5 &86.3 &87.7 &87.8\\
    & AdvPC & 69.1 & 76.9 & 79.2 &74.5 & 74.3 &76.1\\
    & SPSA &76.1 &77.0 &74.4 &74.5 &77.0 &78.9\\
    & Nattack &74.9 &76.5 &73.9 &74.0 &76.3 &77.2\\
\hline
\makecell[c]{$\ell_0$ \\ $\epsilon=200$} & PD & 61.3 &72.1 &73.5 &75.9 &74.1 &74.4\\
\noalign{\global\arrayrulewidth1pt}\hline\noalign{\global\arrayrulewidth0.2pt}
\end{tabular}
  }
\end{table}
\section{Discussion}

Adversarial robustness has been well-established in 2D vision tasks, where Carlini~\etal~\cite{carlini2019evaluating} and many other researchers have devoted significant efforts to set up a rigorous evaluation protocol. In this study, we also emphasize that this evaluation protocol should be followed in 3D point cloud robustness study. Counter-intuitively, we have demonstrated that standard adversarial training is not a good candidate to deliver robustness against strong black-box adversaries because of \textit{gradient obfuscation}. We propose \sysname~as an adversarial purification strategy to mitigate the robustness loss in the 3D space. We would like to clarify that almost all purification methods (including~\sysname) still depend on \textit{gradient obfuscation}. However, we argue that proper usage of \textit{gradient obfuscation} could still serve as a good defense, as long as the obfuscation is sophisticated enough. The multi-step purification in diffusion models adds extremely high-level randomness that EOT~\cite{pmlr-v80-athalye18b} and BPDA~\cite{athalye2018obfuscated} attacks are hard to model. Therefore, we believe our extensive evaluation reveals the actual robustness of \sysname.

\textbf{Broader Impacts and Limitations}. Mitigation solutions to adversarial attacks are critical and essential for modern machine learning systems. Given that 3D point cloud is heavily adopted in safety-critical applications, we believe our study is valuable in demonstrating the vulnerabilities of existing SOTA defenses. \sysname~also. On the other hand, diffusion models needs multiple steps in the reverse process to recover the point cloud and hinder adaptive attacks, which will incur additional computational overhead. \sysname~also limits itself to empirical robustness without theoretical guarantees. Therefore, we cannot exclude possibilities that \sysname~could be broken by future stronger attacks. We plan to include certified defense (\ie randomized smoothing~\cite{cohen2019certified}) into our framework in the near future. 

\section{Conclusion}

In this paper, we propose \sysname, an adversarial purification method against attacks on 3D point cloud recognition. We have first demonstrated that adversarial training and prior purification methods are actually vulnerable to strong attacks. We further leverage extensive evaluation to validate that \sysname~outperforms existing SOTA methods by a significant margin in robust accuracy.  

\bibliographystyle{abbrv}

\bibliography{reference.bib}

\end{document}